\documentclass[journal]{IEEEtran}

\usepackage{cite}
\usepackage{amsmath,amssymb,amsfonts}
\usepackage{algorithmic}
\usepackage{graphicx}
\usepackage{textcomp}

\usepackage{bm}
\usepackage{epsfig}
\usepackage{subfigure}
\usepackage{multirow}
\usepackage{url}
\usepackage{epstopdf}
\usepackage{xcolor}
\usepackage{colortbl}
\usepackage{color, soul}
\soulregister\cite7 
\soulregister\citep7 
\soulregister\citet7 
\soulregister\ref7 
\soulregister\pageref7

\begin{document}

\title{Unifying Relational Sentence Generation and Retrieval for Medical Image Report Composition}
\author{Fuyu~Wang,
        Xiaodan~Liang,
        Lin~Xu,
        and Liang~Lin, \IEEEmembership{Senior Member,~IEEE}
\thanks{Manuscript received March 25, 2020; revised August 31, 2020; accepted September 18, 2020. We thank all anonymous reviewers for their constructive comments. This work was supported in part by National Key RD Program of China under Grant No. 2018AAA0100300, National Natural Science Foundation of China (NSFC) under Grant No.U19A2073 and No.61976233, Guangdong Province Basic and Applied Basic Research (Regional Joint Fund-Key) Grant No.2019B1515120039, Nature Science Foundation of Shenzhen Under Grant No. 2019191361, Zhijiang Lab’s Open Fund (No. 2020AA3AB14). 
All authors are with Sun Yat-sen University, China. Corresponding Author: Liang Lin, Email: linliang@ieee.org.

Fuyu Wang is with the School of Data and Computer Science, Sun Yat-sen University, China (e-mail: wangfy8@mail2.sysu.edu.cn).

Xiaodan Liang is with the School of Intelligent Systems Engineering, Sun Yat-sen University, China (e-mail: xdliang328@gmail.com).

Lin Xu is with the School of Data and Computer Science, Sun Yat-sen University, China (e-mail: cathyxl2016@gmail.com).

Liang Lin is with the School of Data and Computer Science, Sun Yat-sen University, China (e-mail: linliang@ieee.org).
}}

\maketitle

\begin{abstract}
Beyond generating long and topic-coherent paragraphs in traditional captioning tasks, the medical image report composition task poses more task-oriented challenges by requiring both the highly-accurate medical term diagnosis and multiple heterogeneous forms of information including impression and findings.
Current methods often generate the most common sentences due to dataset bias for individual case, regardless of whether the sentences properly capture key entities and relationships. Such limitations severely hinder their applicability and generalization capability in medical report composition where the most critical sentences lie in the descriptions of abnormal diseases that are relatively rare. Moreover, some medical terms appearing in one report are often entangled with each other and co-occurred, e.g. symptoms associated with a specific disease. To enforce the semantic consistency of medical terms to be incorporated into the final reports and encourage the sentence generation for rare abnormal descriptions, we propose a novel framework that unifies template retrieval and sentence generation to handle both common and rare abnormality while ensuring the semantic-coherency among the detected medical terms. Specifically, our approach exploits hybrid-knowledge co-reasoning: i) explicit relationships among all abnormal medical terms to induce the visual attention learning and topic representation encoding for better topic-oriented symptoms descriptions; ii) adaptive generation mode that changes between the template retrieval and sentence generation according to a contextual topic encoder. Experimental results on two medical report benchmarks demonstrate the superiority of the proposed framework in terms of both human and metrics evaluation.
\end{abstract}

\begin{IEEEkeywords}
Multi-modal data processing, Computer-aided diagnosis, Deep learning, Natural language processing
\end{IEEEkeywords}

\section{Introduction}
\label{sec:introduction}
\IEEEPARstart{A}{utomatic} generation of medical image reports has recently attracted increasing research interests \cite{li2018hybrid,jing2017automatic,Dai2019Healthcare}, which has a significant potential to simplify the diagnostic procedure and reduce the burden of physicians. Besides the difficulties shared with captioning and visual question answering (VQA) (e.g. fine-grained visual processing and reasoning, bridging visual and linguistic modalities), the medical report composition must have a plausible logic and consistent topics to complete a long narrative consisting of multiple sentences or paragraphs. Moreover, a task-oriented challenge requires predicting not only the highly accurate medical term diagnosis but also the heterogeneous forms of information including impression and findings.

\begin{figure}
   \centering
   \includegraphics[width=0.5\textwidth]{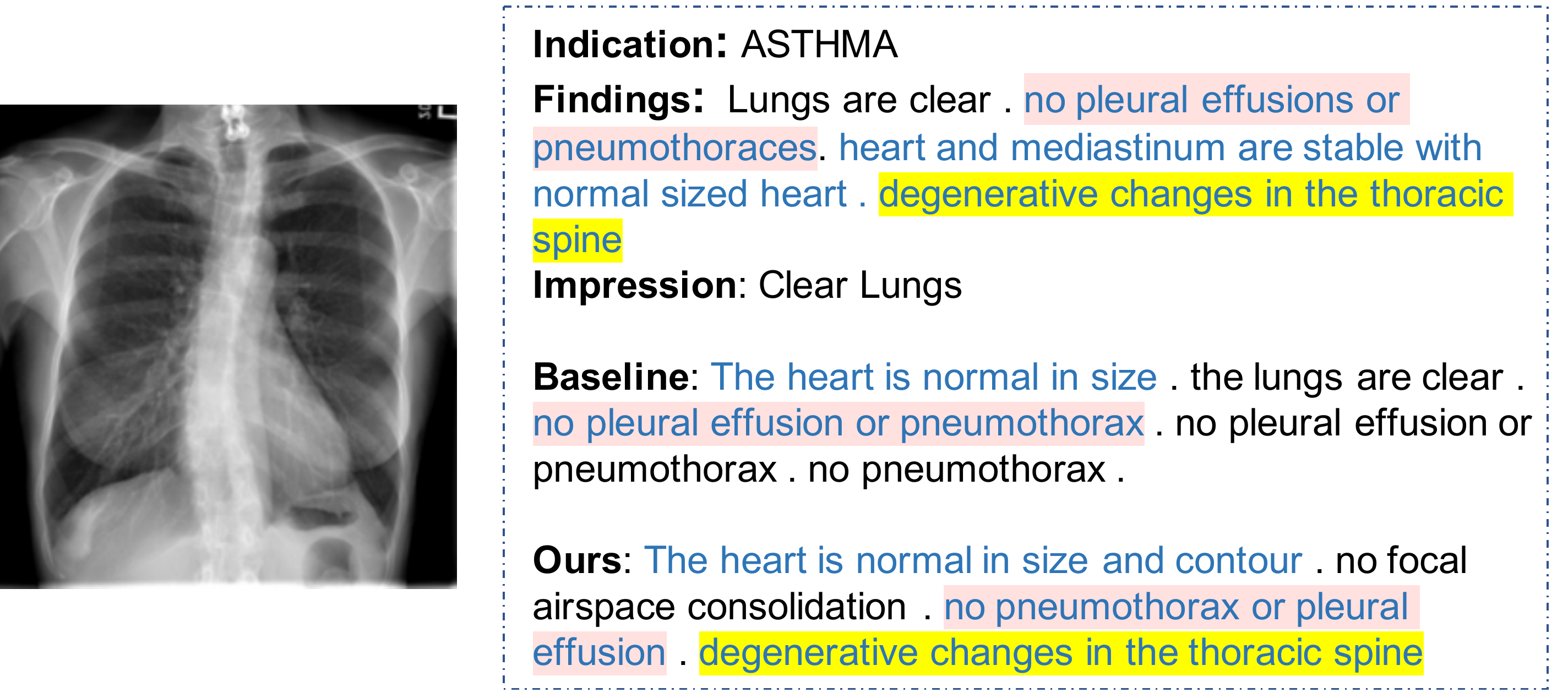}
    \caption{An example of a chest x-ray image and its medical report results of radiologist, baseline and our Relation-paraNet. Superior to the baseline method that produces only normal and repeated parts descriptions, our method can generate degenerative changes in the thoracic spine (abnormality) owes to the relational graph guidance and abnormal template generation method.}
   \label{fig:example}
\end{figure}

Deep neural network architectures \cite{huang2017densely,simonyan2014very,Wang2019LungCancer,Chen2017Ultrasound}, sequence-to-sequence models \cite{gu2016incorporating,wang2018reinforced} and visual attention mechanisms \cite{anderson2018bottom,jing2017automatic,kim2018multimodal} have been widely adopted in both image captioning and VQA, which improve performance by learning to focus on the salient regions of the image. However, without other prior knowledge on the visual content, such computed visual attention may concentrate on irrelevant regions. Furthermore, few methods have considered key entities, topic relationships and paragraph consistency, which are most likely to generate similar sentences in one report or similar reports for different medical images, due to the dataset bias.

In general, radiologists first check a patient's images for peculiar regions, think of correlations among prominent symptoms, then write sentences according to the keywords following certain patterns for the normal cases and adjust statements for the specific cases. In this paper, we adopted a similar methodology and proposed a Unifying Retrieval and Relational-topic sentence Generation framework, named Relation-paraNet, which incorporates the semantic consistency of medical terms into the reports and encourages the sentence generation for rare abnormal descriptions. Specifically, our Relation-paraNet exploits hybrid-knowledge co-reasoning in two ways. First, we explore explicit relationships among all abnormal medical terms to induce the visual attention learning and topic representation encoding for better topic-oriented symptom descriptions. We pay attention to the abnormal medical terms that reflect the keywords of the reports and introduce the relational-topic to guide the sentence generation. We further excavate the semantic consistency among all abnormal keywords and conduct abnormality classification to induce visual attention learning. On the other hand, to generate better topic-oriented symptom descriptions, we integrate visual features and abnormality relations for topic representation, which is essential to depict the principal idea of the independent sentence in the report.

Furthermore, inspired by the fact that radiologists often write reports based on templates, we introduce an Adaptive Generator that changes between the template retrieval and sentence generation according to a contextual topic encoder. We employ a retrieval classification module to decide the choice for either automatically generating sentences or retrieving specific sentences from the template database. The template database is based on human prior knowledge collected from available medical reports. To enable effective and robust report composition, we take the integral generated sentences into concern and encode such paragraph information back into the network to produce the next sentence. Experimental results show that by unifying retrieval paragraph and relational topic, our Relation-paraNet is able to generate more accurate and professional medical reports.

Our contributions are summarized in the following aspects.
\begin{itemize}
\item Aiming at resolving the task-oriented challenges of medical report composition, we make the first attempt to incorporate the semantic consistency of medical terms into the final reports and encourage the precise sentence generation for rare abnormal descriptions.
\item We introduce a Unifying Retrieval and Relational-topic driven Generation framework, named Relation-paraNet, which integrates a Relational-Topic Encoder to learn explicit semantic consistency among medical terms and an Adaptive Generator to change between the template retrieval and sentence generation for more natural medical report composition.
\item Our method outperforms state-of-the-art works on two medical report datasets and achieves an appealing performance under human evaluation.
\end{itemize}

The rest of this paper is organized as follows. Section II
presents a brief review of related work. Section III introduces
the pipeline of our method, including the
model formulation and the optimization method. The
experimental results, comparisons, and component analysis
are presented in Section IV. Section V concludes the paper with a discussion of future work.

\begin{figure*}
   \centering
   \includegraphics[width=1.05\textwidth]{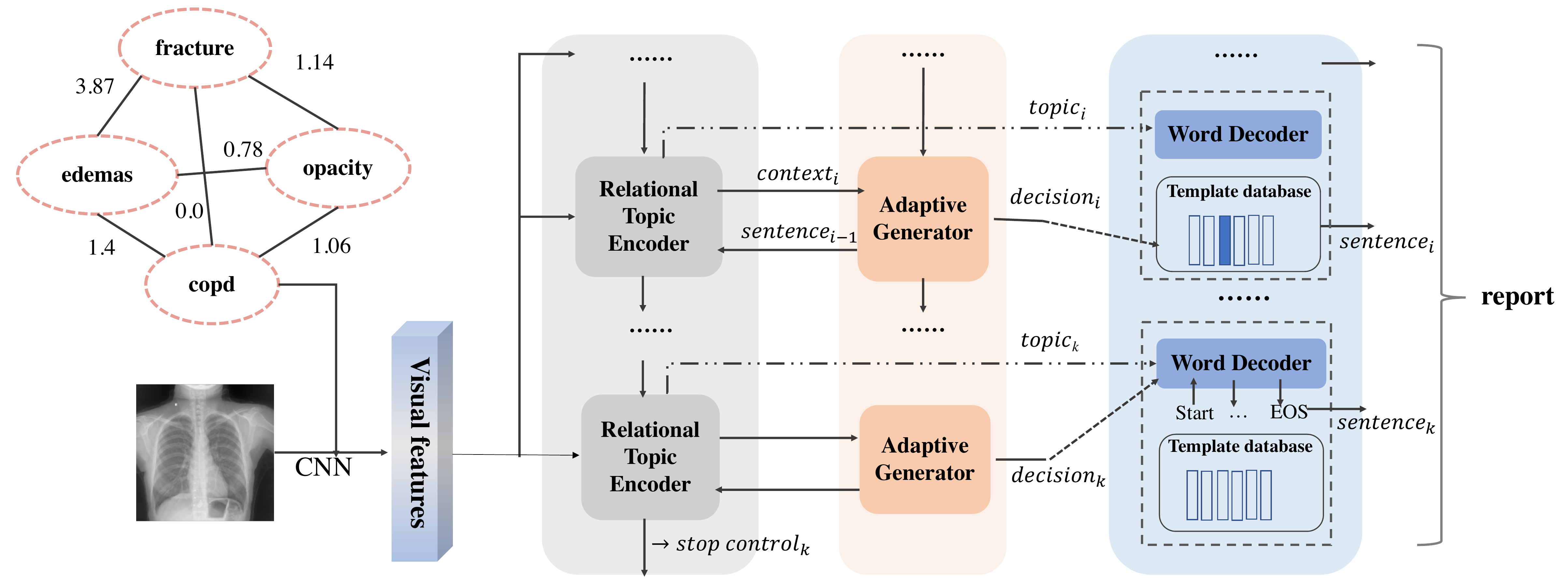}
    \caption{An overview of our Relation-paraNet, which consists of two cooperative modules, the Relational-Topic Encoder and the Adaptive Generator. Deep CNN architecture is used to learn visual features and explore the semantic consistency of medical terms. The Relational-Topic Encoder produces topic vectors and the Adaptive Generator retrieves a template or generates a sentence.}
   \label{fig:overall}
\end{figure*}

\section{Related Work}
\label{related_work}

\textbf{Visual Captioning} is a challenging cognitive task that requires a simultaneous understanding of both natural language and visual information. Early pioneering methods based on deep learning \cite{krizhevsky2012imagenet} have employed CNN \cite{simonyan2014very, he2016deep} and RNN \cite{hochreiter1997long, cho2014learning} to generate syntactical sentences for images or videos. For example, Karpathy \textit{et al.} \cite{karpathy2015deep} proposed an alignment model to generate descriptions of image regions by incorporating R-CNN \cite{girshick2014rich} features into RNN.
Vinyals \textit{et al.} \cite{vinyals2015show} devised an end-to-end captioning system by utilizing LSTM to maintain information in memory, which is improved via a spatial attention mechanism to automatically localize related regions in \cite{xu2015show}. 

Based on the above prevailing frameworks, recent approaches \cite{xu2018dual, li2019know, gao2017video, zhang2018high, xiao2019deep, chen2019generating, Dai2019Healthcare} towards different ambitions have been proposed. Xu \textit{et al.} \cite{xu2018dual} designed a dual-stream RNN architecture to exploit both visual and semantic features jointly.
To address the issue that the image is segmented by CNN to the fixed resolution grid at a coarse level, Zhang \textit{et al.} \cite{zhang2018high} equipped the image captioning model with fine-grained and semantic-guided visual attention. Yao \textit{et al.} \cite{yao2019hierarchy} introduced a hierarchy from instance level and region level to delve into a thorough image understanding. Then a HIerarchy Parsing architecture is used to integrate the hierarchical structure into image encoder.
Furthermore, to learn structured semantic knowledge in scene graphs,
Li \textit{et al.} \cite{li2019know} first extracted triples from scene graphs and encoded them into semantic vectors, then devised a hierarchical-attention-based module to focus on salient visual and semantic features.
Since the capacity of a single-LSTM network is limited for image captioning,
Xiao \textit{et al.} \cite{xiao2019deep} developed a deep hierarchical structure to fuse multi-level semantics of vision and language by increasing the vertical depth of the encoder-decoder.

\textbf{Medical Report Composition} is one of the challenging research topics in machine learning for healthcare \cite{li2018hybrid, jing2017automatic,Dai2019Healthcare}. Different from image captioning, this task requires generating multiple sentences and puts forward higher requirements on content selection, relation generation, and content fluency.
Jing \textit{et al.} \cite{jing2017automatic} proposed a co-attention mechanism to focus on abnormal regions and a hierarchical LSTM (Sentence LSTM and Word LSTM) like \cite{krause2017hierarchical} to generate long paragraphs. 
To incorporate patient background information, Huang \textit{et al.} \cite{huang2019multi} proposed a hierarchical model with multi-attention mechanism. The patient’s background information is encoded and then added to the pretrained vanilla word embedding.
Due to relatively rare and remarkably diverse abnormal findings, \cite{jing2017automatic} fails to detect abnormalities and tends to generate trivial descriptions. Li \textit{et al.} \cite{li2019knowledge} directly selected sentences from the template database via Graph Transformer \cite{velickovic2018graph}. 
Recently, there are some approaches combining traditional template-based and newest generation-based methods for sequence generation\cite{cao2018retrieve, gu2016incorporating, chen2018fast}. Cao \textit{et al.} \cite{cao2018retrieve} used existing summaries as soft templates to guide the generative model, jointly applying template retrieval, template re-ranking and template-aware summary generation.
Differing from above methods failing to incorporate semantic consistency with medical abnormalities for sequence generation, our method unifies template retrieval and sentence generation to handle both common and rare abnormality while ensuring the semantic-coherency among the detected medical terms.
Similar to us, Li \textit{et al.} \cite{li2018hybrid} combined retrieval-based and generation-based methods via reinforcement learning, whereas our generated descriptions are more consistent with detected abnormalities. Furthermore, Nie \textit{et al.} \cite{nie2014bridging} proposed a novel scheme that combines local mining and global learning approaches to bridge the vocabulary gap between health seekers and providers.

\textbf{Attention Mechanism} tries to learn different attention weights for input text or image and focus on informative region automatically \cite{bahdanau2014neural, vaswani2017attention}, which can be also applied to visual captioning \cite{anderson2018bottom, xu2015show, hori2017attention, jin2015aligning, xu2017learning} and report composition \cite{li2019knowledge}. For instance, Anderson \textit{et al.} \cite{anderson2018bottom} combined bottom-up and top-down attention mechanism in which the latter mechanism can calculate attention at the level of objects extracted by the former mechanism. Hori \textit{et al.} \cite{hori2017attention} introduced a multimodal attention model that selectively focus on specific information across different modalities for video description. Gao \textit{et al.} \cite{gao2017video} developed an attention-based framework to exploit the correlations between sentence semantics and visual content by mapping the features of them into a joint space. And Pan \textit{et al.} \cite{pan2020x} devised an X-Linear attention block to capture the $2^{nd}$ order feature interaction in between, and measures both spatial and channel-wise attention distributions. To make the decoder determine the relevance between the attended vector and the given attention query, Huang \textit{et al.} \cite{huang2019attention}  proposed an Attention on Attention module from which the useful knowledge was produced.

\textbf{Multi-label Classification}
Multi-label classification is a challenging task that attracts increasing attention \cite{xue2011correlative,wang2016cnn,guillaumin2009tagprop,gong2013deep,wei2014cnn}.  Recent approaches, which are optimized with the ranking loss \cite{gong2013deep} or the cross-entropy loss \cite{guillaumin2009tagprop} incorporate deep convolutional neural network into multi-label classification and achieve superior performance. However, this task exposes strong label co-occurrence dependencies according to \cite{xue2011correlative}. Thus Wang \textit{et al.} \cite{wang2016cnn} proposed CNN-RNN framework learns a joint image-label embedding to exploit label dependencies in an image.  Furthermore, Feature Attention Network(FAN) \cite{yan2019multi}, which builds the top-down feature fusion mechanism and learn the correlations among convolutional features, is designed to tackle object scale inconsistent and label dependencies. Chen \textit{et al.} \cite{chen2019deep} introduced a two-stage method that trained the network to predict conditional probability directly and then refined it with unconditional probabilities which formulated from a numerically stable and principled loss function. Different from the above methods, we construct a relationship constraint loss function to learn label co-occurrence dependencies directly.

\section{Method}

As shown in Figure \ref{fig:example}, a complete diagnostic report for a medical image is comprised of both text descriptions and lists of medical terms. Inspired by \cite{krause2017hierarchical, li2018hybrid}, we formulate the generation in a hierarchical framework illustrated in Figure \ref{fig:overall}. A medical image is first fed into a deep CNN architecture to learn visual features. We explore the semantic consistency of medical terms and perform the abnormality classification to encourage the sentence generation of rare abnormal descriptions. To achieve semantic alignment between visual features and report descriptions, the Relational-Topic Encoder and Adaptive Generator integrate the advantages of the bottom-up and top-down attention mechanism. Taking image features $\bm{v}$ and the sentence embedding as input, the Relational-Topic Encoder produces a contextual topic vector $\bm{c}_i$ and a stop control $\bm{z}_k$ to determine the number of sentences. The Adaptive Generator, acting as a gate function, takes the encoded contextual topic vector $\bm{c}_i$ as input to determine the choice to either retrieve a template from the template database or generate a new sentence. In the latter option, the sentence topic vector $\bm{q}_i$ is fed into the Word Decoder to generate words for the corresponding sentence. The proposed model produces two topic vectors, $\bm{c}_i$ and $\bm{q}_i$, respectively.


\subsection{Relational Abnormality Classification}
\label{relational_abnormality}
Previous works \cite{jing2017automatic, anderson2018bottom} have shown that visual attention can perform fairly well for localizing objects and aiding captions. However, visual attention is usually not sufficient to encode high-level semantic information to recognize abnormalities. To this end, we explore the relationship between the medical terms which can be cooperative with image features for the robust relational topic generation.

We treat the abnormality classification as a multi-label image classification task. Specifically, given a medical image $I$, we first extract its feature maps through a deep CNN and apply a fully connected layer to compute a distribution over all of the abnormal medical terms. To solve multi-label classification, we calculate binary cross entropy for each category. Moreover, according to \cite{xue2011correlative,wang2020multi}, labels co-occur in images with priors. And relationships among all abnormal medical terms are also crucial for classification as some terms are often entangled with others and co-occurred. For example, ``copd'' and ``pulmonary disease chronic obstructive'' often appear together. Therefore, the predicted scores of these two abnormalities should be closer. To exploit these explicit relationships, beyond average binary cross entropy loss, we add another relationship constraint loss. Finally, the abnormality multi-label classification loss is composed of two terms, which is formulated as:
\begin{equation}
\label{eq:classification_loss}
\begin{aligned}
   \mathcal{L}_{cls} = &\frac{1}{M}\sum_{i}^{M}{ [y_i\log{a_i} + (1 - y_i)\log{(1 - a_i)}] } \\
   & + \frac{1}{R^{\ast}}\sum_{i}^{M}{\sum_{j}^{M}{(a_{i} - a_{j})^2 * r(i, j)}},
\end{aligned}
\end{equation}
where the first term is an average binary cross-entropy loss for each category, $a_{i}$ denotes probability of abnormal medical term $i$ and $y_i$ is the ground truth label; the second term is the relationship constraint loss, $r(i, j)$ represents the relevance between two abnormal medical terms and $R^{\ast}$ denotes the number of non-zeros in relationship matrix $r$. For a pair of abnormality probabilities, namely $a_{i}$ and $a_{j}$, if $r(i, j)$ is larger, the relationship constraint loss can guide $a_{i}$ and $a_{j}$ to be closer. Similarly, $a_{i}$ and $a_{j}$ do not influence each other when $r(i, j)$ is smaller.

The static relationship matrix is computed according to the frequency of co-occurrences for each pair of abnormalities all over the training set \cite{fang2017object}, which is defined as follows:
\begin{align}
   \label{eq:relation_matrix_comput}
   r(i,j) = \max(\log{\frac{f(i, j)F}{f(i)f(j)}}, 0),
\end{align}
where $f(i, j)$ denotes the frequency of co-occurrences, $f(i)$ is the frequency of abnormality $i$, and $F$ denotes the total count of abnormalities. Here we do not consider the extreme case when $f(i,j)=f(i)=f(j)$, and we only make the statistics for a rough prior knowledge.

\subsection{Relational-Topic Encoder}
\begin{figure}
   \centering
   \includegraphics[width=0.5\textwidth]{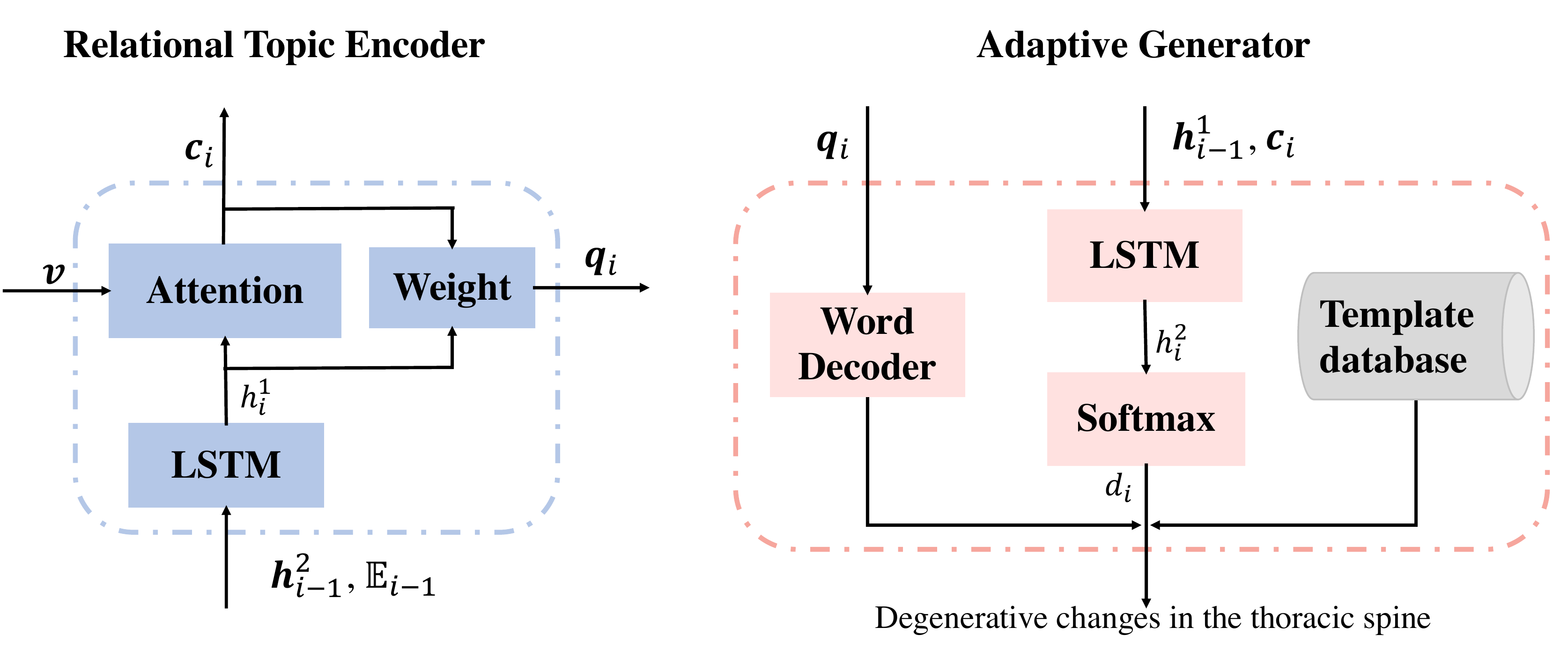}
   \caption{Detailed illustration of the Relational-Topic Encoder and Adaptive Generator. Composed of an LSTM layer, an attention layer and a fully connected layer, the Relational-Topic Encoder takes image features $\bm{v}$ and previous generated sentence embedding as inputs, and then outputs a context vector $\bm{c}_i$ and topic vector $\bm{q}_i$, which are put into the Adaptive Generator to produce a sentence via the Word Decoder or the Template database.}
   \label{fig:topic_template_detail}
\end{figure}

Sentences in the report are based on the image as well as the previous generated sentence. In order to utilize the relationships among abnormal medical terms and adjacent sentences, we propose a Relational-Topic Encoder to generate more discriminative topics, as illustrated in Figure \ref{fig:topic_template_detail}. Our Relational-Topic Encoder has two critical inputs. The first one is the image features which are enhanced by the abnormal medical terms learning. Its visual information can help the encoder focus on the salient regions. Furthermore, we also feed the embedding vector of the previous sentence into the decoder, which imposes the encoder to memorize the previous topic and sentence features.
Formally, the Relational-Topic Encoder is composed of an LSTM layer and an attention \cite{xu2015show} layer. At each time step, the LSTM layer concatenates the previous state of the Adaptive Generator $\bm{h}_{i-1}^{2}$ and embedding vector of the previous generated sentence $\mathbb{E}_{i-1}$ to form the input vector $\bm{x}_{i}^{1}$:
\begin{equation}
\bm{x}_{i}^{1} = [\bm{h}_{i-1}^{2}; \mathbb{E}_{i-1}],
\end{equation}

Supposing a sentence consists of $n$ tokens, a bi-directional LSTM is adapted to encode the sentence as a 2-D matrix \cite{britz2017massive}.
The proposed Relational-Topic Encoder generates a hidden state $\bm{h}_{i}^1$ by an {LSTM} as follows,
\begin{equation}
\bm{h}_{i}^{1} = {LSTM}(\bm{x}_{i}^{1}, \bm{h}_{i-1}^{1}).
\label{butd1}
\end{equation}
This hidden state $\bm{h}_{i}^1$ is used to produce three signals. First, $\bm{h}_{i}^1$ is linearly projected into a $\textit{stop control}$ $\bm{z}_i$, which determines whether to stop the topic generation process or not,
\begin{equation}
\bm{z}_{i} = {Sigmoid}(\bm{W}_{z}\bm{h}_{i}^1 + \bm{b}_{z}),
\end{equation}
where $\bm{W}_{z}$ an $\bm{b}_{z}$ are trainable parameters. 
Second, given a hidden state $\bm{h}_{i}^1$ together with image features $\bm{v}$, an attended context vector $\bm{c}_{i}$ is produced with an attention operation $\textbf{att}$,
\begin{equation}
\bm{c}_{i} = \textbf{att}(\bm{h}_{i}^{1}, \bm{v}).
\end{equation}
Third, the hidden state $\bm{h}_{i}^1$ and the context vector $\bm{c}_{i}$ are fed into a fully-connected layer with weights $\bm{W}_{q,h}$ and $\bm{W}_{q,c}$ to produce the topic vector $\bm{q}_{i}$,
\begin{equation}
\bm{q}_{i} = \bm{W}_{q,h}\bm{h}_{i}^{1} + \bm{W}_{q,c}\bm{c}_{i}.
\end{equation}

In this way, a relational sentence topic vector $\bm{q}_{i}$ is generated for the word decoder to predict words sequentially. Simultaneously, an encoded contextual topic vector $\bm{c}_{i}$ is produced to retrieve templates adaptively.

\subsection{Adaptive Generator}
As the frequency of normal sentences is much higher than that of abnormal sentences, state-of-the-art methods on report composition \cite{jing2017automatic} tend to generate normal sentences such as ``The heart is normal in size'', ``The lungs are clear'' and ``No acute bony abnormality''. As for the abnormal sentences like ``Scattered thoracic spine spurring'', these models cannot write accurately due to the dataset bias. Different from the previous methods, our proposed Adaptive Generator produces the abnormal sentences from template database and employs the Word Decoder for generating normal sentences, which combines retrieval and generation for automatic report composition, as shown in Figure \ref{fig:topic_template_detail}. During training, if the module needs to select a template, the Word Decoder will be masked and vice versa.

In particular, the Adaptive Generator takes encoded contextual topic vector $\bm{c}_{i}$ as input to determine the choice that performances either  template retrieval or sentence writing for the current sentence generation. It consists of an LSTM layer and a Softmax classifier. Given a hidden state $\bm{h}_{i}^1$ from the Relational-Topic Encoder and contextual topic vector $\bm{c}_{i}$ predicted by the attention layer, the Adaptive Generator produces the adaptive decision $\bm{d}_i$ for sentence $i$. Since sentence writing is also one of the decisions, the size of the template database is $N$ and the decision space is $N+1$. The formulation can be summarized as:
\vspace{-2mm}   
\begin{equation}
\label{eq:adaptive_generation}
\begin{aligned}
   \bm{x}_{i}^{2} &= [\bm{h}_{i}^{1}, \bm{c}_{i}], \\
   \bm{h}_{i}^{2} &= LSTM(\bm{x}_{i}^{2}, \bm{h}_{i-1}^{2}), \\
   \bm{d}_{i} &= {Softmax}(\bm{W}_{d}\bm{h}_{i}^2 + \bm{b}_{d}).
\vspace{-4mm}
\end{aligned}
\end{equation}
\vspace{-2mm}   


From the above Equation \ref{butd1} and \ref{eq:adaptive_generation}, in which two LSTM layers are used to selectively attend to spatial image features, we can see that the bottom-up and top-down attention mechanism \cite{anderson2018bottom} is integrated into our Relational-Topic Encoder and Adaptive Generator. 

As shown in Figure \ref{fig:overall}, the relational topic vector $\bm{q}_{i}$ represents the overall feature of the generated sentence. The topic vector and a special \textit{START} token are regarded as the inputs to the Word Decoder following the previous work \cite{jing2017automatic, li2018hybrid}. The subsequent inputs are the word embedding sequence. 
\vspace{-2mm} 
\begin{equation}
   \bm{p}_{w} = {Softmax}(\bm{W}_{w}\bm{h}_{w} + \bm{b}_{w}).
\vspace{-3mm}   
\end{equation}
\vspace{-3mm}   

For each word, the last hidden state $\bm{h}_{w}$of the Word Decoder is used to predict a distribution over the words in the vocabulary. Finally, all sentences from retrieval or writing are concatenated to form the medical report. 

\subsection{Parameter Learning}
\label{parameter_learning}
Training data consists of tuples $(I, \bm{y}, \bm{r})$, where $I$ is an image, $\bm{y}$ denotes the ground-truth abnormalities and $\bm{r}$ is the ground-truth report description, which has $m$ sentences. To perform adaptive generation, the $i$-th sentence has a template index ($t_i \in [0, N]$) as well as $n$ word indexes ($w_{i_1}, w_{i_2}, ..., w_{i_n}$).

Given a tuple of $(I, \bm{y}, \bm{r})$, we perform multi-label abnormality classification and unroll the Sentence Decoder $m$ timesteps, receiving abnormality distribution $a_i$, stop distribution $\bm{z}_i$ over the $\{\rm CONTINUE, \rm STOP\}$ states, and template distribution $\bm{d}_i$ over the template database. Then we unroll the Word Decoder $n$ timesteps if the adaptive generation mode changes to sentence generation, receiving word probability $\bm{p}_{i,j}$.

Formally, we define a multi-task loss similar to \cite{krause2017hierarchical}:
\vspace{-2mm}   
\begin{equation}
\begin{aligned}
\mathcal{L} &= \mathcal{L}_{cls} + \mathcal{L}_{stop} + \mathcal{L}_{tem} * \mathcal{M}_{tem} + \mathcal{L}_{word} * \mathcal{M}_{word} \\
\mathcal{L}_{stop} &= \frac{1}{m}\sum_{i=1}^m {\rm BCE}(\bm{z}_{i}, \bm{I}[i = m]) \\
\mathcal{L}_{tem} &= \frac{1}{m}\sum_{i=1}^m {\rm CE}(\bm{d}_{i}, t_i) \\
\mathcal{L}_{word} &= \frac{1}{m*n}\sum_{i=1}^{m'}\sum_{j=1}^{n} {\rm CE}(\bm{p}_{i,j}, w_{i,j}),
\label{total_loss}
\end{aligned}
\end{equation}
where $\rm (B)CE$ denotes (binary) cross entropy function, $\mathcal{L}_{cls}$ is the abnormality multi-label classification loss consisting of binary cross-entropy loss and relationship constraint loss defined in Equation \ref{eq:classification_loss}, 
$\mathcal{L}_{stop}$ denotes the stop signal prediction loss, $\mathcal{L}_{tem}$ and $\mathcal{L}_{word}$ are the template classification loss for each sentence and the word prediction loss over word distributions respectively, $m$ denotes the number of sentences that are generated by Word Decoder. If our model needs to select a template, the masks $\mathcal{M}_{tem}$ and $\mathcal{M}_{word}$ are set to $1$ and $0$ respectively, and vice versa. All the components are jointly trained to minimize $\mathcal{L}$.

\begin{table}[t]
    \caption{Abnormality classification comparison.}
    \label{tab:classification_results}
    \centering
    \tiny
    \resizebox{0.9\columnwidth}{!}{
    \begin{tabular}{c|c|c}
    \hline
    Dataset & Method & AUC  \\
    \hline
    \multirow{2}{*}{IU X-Ray} & VGG-19 \cite{simonyan2014very} & 74.2 \\
    ~& \textbf{Relation-paraNet} & \textbf{75.1}  \\
    \hline
    \hline
    \multirow{2}{*}{CX-CHR} & DenseNet-121 \cite{huang2017densely} & 84.8 \\
    ~& \textbf{Relation-paraNet} & \textbf{86.7}  \\
    \hline
    \end{tabular}
    }
    \vspace{-4mm}
\end{table}

\begin{table*}[t]
    \caption{Performance comparison with state-of-the-art methods on both IU X-Ray and CX-CHR dataset. Our model outperforms all existing state-of-the-art approaches by a large margin on automatic evaluation metrics.}
    \label{tab:caption_results}
    \centering
     \resizebox{\textwidth}{!}{
    \begin{tabular}{c|c|*{7}{c}}
    \hline
    Dataset & Model & CIDEr & ROUGE-L & BLEU-1 & BLEU-2 & BLEU-3 & BLEU-4  \\
    \hline
    \multirow{6}{*}{IU X-Ray} & CNN-RNN\cite{vinyals2015show} & 0.294 & 0.307 & 0.216 & 0.214 & 0.087 & 0.066 \\
    ~& LRCN \cite{donahue2015long} & 0.285 & 0.307 & 0.223 & 0.128 & 0.089 & 0.068 \\
    ~& AdaAtt \cite{lu2017knowing} & 0.296 & 0.308 & 0.220 & 0.127 & 0.089 & 0.069 \\
    ~& Att2in \cite{rennie2017self} & 0.297 & 0.307 & 0.224 & 0.129 & 0.089 & 0.068 \\
    ~& Transformer \cite{vaswani2017attention} & 0.255 & 0.371 & 0.363 & 0.231 & 0.155 & 0.107  \\
    ~& BUTD \cite{anderson2018bottom} & 0.312 & 0.368 & 0.398 & 0.256 & 0.174 & 0.124 \\
    ~& AoA \cite{huang2019attention} & 0.292 & 0.355 & 0.204 & 0.119 & 0.085 & 0.062 \\
    ~& CoAtt \cite{jing2017automatic} & 0.277 & 0.369 & 0.455 & 0.288 & 0.205 & {0.154}  \\
    ~& HRGR-Agent \cite{li2018hybrid} & \textbf{0.343} & 0.322 & 0.438 & 0.298 & 0.208 & 0.151 \\
    ~& \textbf{Relation-paraNet(VGG-19)} & 0.317 & \textbf{0.372} & \textbf{0.505} & 0.329 & 0.230 & 0.168  \\
    \hline
    ~& KERP \cite{li2019knowledge} & 0.280 & 0.339 & 0.482 & 0.325 & 0.226 & 0.162 \\
    ~& \textbf{Relation-paraNet(DenseNet-121)} & 0.331 & 0.360 & 0.503 & \textbf{0.333} & \textbf{0.236} & \textbf{0.175}  \\

    \hline
    \hline
    \multirow{6}{*}{CX-CHR} & CNN-RNN\cite{vinyals2015show} & 1.580 & 0.577 & 0.590 & 0.506 & 0.450 & 0.411 \\
    ~& LRCN \cite{donahue2015long} & 1.588 & 0.577 & 0.593 & 0.508 & 0.452 & 0.413 \\
    ~& AdaAtt \cite{lu2017knowing} & 1.568 & 0.575 & 0.588 & 0.503 & 0.446 & 0.409 \\
    ~& Att2in \cite{rennie2017self} & 1.566 & 0.576 & 0.587 & 0.503 & 0.446 & 0.408 \\
    ~& Transformer \cite{vaswani2017attention} & 2.721 & 0.605 & 0.590 & 0.527 & 0.484 & 0.453 \\
    ~& BUTD \cite{anderson2018bottom} & 2.516 & 0.660 & 0.642 & 0.562 & 0.506 & 0.462 \\
    ~& AoA \cite{huang2019attention} & 1.774 & 0.617 & 0.606 & 0.516 & 0.455 & 0.408 \\
    ~& CoAtt \cite{jing2017automatic} & 2.735 & 0.645 & 0.647 & 0.575 & 0.525 & 0.487 \\
    ~& HRGR-Agent \cite{li2018hybrid} & 2.895 & 0.612 & 0.673 & 0.587 & 0.530 & 0.486  \\
    ~& KERP \cite{li2019knowledge} & 2.850 & 0.618 & 0.673 & 0.588 & 0.532 & 0.473 \\
   ~& \textbf{Relation-paraNet(DenseNet-121)} & \textbf{3.249} & \textbf{0.675} & \textbf{0.711} & \textbf{0.637} & \textbf{0.586} & \textbf{0.548}  \\
    \hline
    \end{tabular}
    }
    \vspace{-4mm}
\end{table*}

\section{Experiments and Analysis}

To evaluate the effectiveness of the model, we conduct experiments on two medical image report datasets.

\textbf{IU X-Ray}. Indiana University Chest X-Ray Collection (IU X-Ray) \cite{demner2015preparing} is a public collection that contains 7470 pairs of images and diagnostic reports. Each report is comprised of impression, findings, tags and indication, which properly meets the requirements of our task. Every token is converted to lower-case, and filtered out with a frequency less than 3, which results in 1185 unique words covering over 99.0\% word occurrences in the corpus. To perform subsequent tasks, we screen tags with a frequency greater than 30 and abnormal sentences with a frequency greater than 2, which produces 80 abnormalities and 80 templates.

\textbf{CX-CHR} is a private collection of chest X-ray images with corresponding Chinese reports for health checking, consists of 35,609 patients and 45,598 images.
There are 33236 patient samples in total, covering over 93\% of the dataset. Different from IU X-Ray, CX-CHR needs to tokenize by Jieba firstly. Then we filter out tokens with the frequency less than 3, tags with the frequency less than 30 and abnormal sentences with the frequency less than 2. Finally, we obtain 1233 unique tokens, 155 abnormalities and 287 templates for CX-CHR dataset. Following \cite{li2018hybrid}, as the ``findings'' section contains the radiological observations and detailed description of the patient¡¯s information, we only consider ``findings" section as the target captions. To protect the privacy of patients, CX-CHR dataset is not publicly available, but the researchers can apply for academic usage after signing the confidentiality agreement.

Since the two datasets are collected from two different hospitals, there exist variance among reports and abnormalities. In this paper, the abnormal sentences are regarded as templates. And the ratio of template retrieval and sentence generation is 0.26 for IU X-Ray(English) dataset. As for CX-CHR(Chinese) dataset, the ratio is 0.16, since the CX-CHR dataset contains more normal sentences than the IU X-Ray dataset.
On both datasets, we use the same dataset split from \cite{li2019knowledge} for a fair comparison. There is no overlap between patients in different sets.

\begin{table*}[t]
    \caption{Comparisons of different components of our Relation-paraNet on both IU X-Ray and CX-CHR dataset.}
    \label{tab:ablation_results}
    \centering
    \resizebox{\textwidth}{!}{
    \begin{tabular}{c|c|*{6}{c}}
    \hline
    Dataset & Model & CIDEr & ROUGE-L & BLEU-1 & BLEU-2 & BLEU-3 & BLEU-4 \\
    \hline
    \multirow{6}{*}{IU X-Ray} ~& w/o Sentence relation & 0.249 & 0.328 & 0.385 & 0.235 & 0.155 & 0.108 \\
    ~& Normal templates & 0.144 & 0.282 & 0.292 & 0.183 & 0.128 & 0.094 \\
    ~& w/o Abnormality relation & 0.317 & 0.330 & 0.355 & 0.217 & 0.150 & 0.111  \\
    ~& \textbf{Relation-paraNet(VGG-19)} & \textbf{0.317} & \textbf{0.372} & \textbf{0.505} & \textbf{0.329} & \textbf{0.230} & \textbf{0.168} \\
    \hline
    \hline
    \multirow{6}{*}{CX-CHR} ~& w/o Sentence relation & 2.996 & 0.655 & 0.667 & 0.597 & 0.548 & 0.510 \\
    ~& Normal templates  & 2.855 & 0.650 & 0.700 & 0.611 & 0.550 & 0.505 \\
    ~& w/o Abnormality relation & 3.064 & 0.657 & 0.711 & 0.636 & 0.586 & 0.547  \\
    ~& \textbf{Relation-paraNet(DenseNet-121)} & \textbf{3.249} & \textbf{0.675} & \textbf{0.711} & \textbf{0.637} & \textbf{0.586} & \textbf{0.548}   \\
    \hline
    \end{tabular}
    }
    \vspace{-4mm}
\end{table*}

\subsection{Implementation Details}
\textbf{Training.}
For two medical datasets, we set the same hyper-parameters, due to our robust captioning network. Our model accepts a $224\times224$ image as input and yields $(H, W, C)$ feature maps from the last convolution layer. $(H,W,C)$ is $(14,14, 512)$ and $(7,7,1024)$ when using VGG-19 \cite{simonyan2014very} or DenseNet-121 \cite{huang2017densely}, respectively. We use 512 as the dimension of all hidden states, topic vectors, word embedding and so on. To minimize $\mathcal{L}$ in Equation \ref{total_loss}, we train on a single GPU with an initial learning rate of $5\times10^{-4}$ and a batch size of 16 for 30 epochs. We employ ADAM \cite{kingma2014adam} optimizer and decrease the learning rate by the factor of 0.8 every three epochs, following \cite{rennie2017self}. 

\textbf{Inference.}
We execute our model to select a template or write sentence through Word Decoder at each time step until a stop control appears or the number of generated sentences reaches the maximum value $S_{Max}$. Here, we terminate the process as long as the score of stop control exceeds a threshold of 0.5. In detail, predicting the template index $``\textbf{0}"$ means we should believe the written sentence, otherwise, copy the corresponding template to the report directly. In the same way, we stop Word Decoder if we predict a \textit{END} token or obtain $W_{Max}$ words. We set $S_{Max}$ and $W_{Max}$ as 10 and 15 for IU X-Ray, 24 and 18 for CX-CHR dataset, respectively.

\subsection{Experimental Results}
\label{experimental_results}
\textbf{Evaluation metrics.}
We conduct experiments to evaluate our method from three aspects: 1) Area Under the Curve (AUC) measures the performance of abnormality classification; 2) Automatic evaluation metrics consist of BLEU \cite{papineni2002bleu}, ROUGE \cite{lin2004rouge} and CIDEr \cite{vedantam2015cider}. BLEU is defined as the geometric mean of n-gram precision scores multiplied by a brevity penalty for short sentences. ROUGE-L basically measures the longest common subsequences between a pair of sentences and CIDEr measures the consensus between candidate image description and the reference sentences; 3) Human evaluation is performed by the radiologists with professional experiments. We invite 5 doctors to select the better generated reports between the baseline method {CoAtt} \cite{jing2017automatic} and ours' proposed model. The selection criteria consists of abnormal findings, language fluency, and content coverage.

 \textbf{Baselines.}
 We compare our Relation-paraNet with 4 state-of-the-art image captioning methods. They are CNN-RNN \cite{vinyals2015show}, LRCN \cite{donahue2015long}, AdaAtt \cite{lu2017knowing}, and Att2in \cite{rennie2017self} respectively. Besides the above methods, we also compare with state-of-the-art methods of medical report composition, including AoA \cite{huang2019attention}, Transformer \cite{vaswani2017attention}, BUTD \cite{anderson2018bottom}, CoAtt \cite{jing2017automatic}, HRGR-Agent \cite{li2018hybrid}, and KERP \cite{li2019knowledge}. For the fair comparison, we employ VGG-19 \cite{simonyan2014very} and DenseNet121 \cite{huang2017densely} to extract visual features on IU X-ray and CX-CHR dataset respectively, in step with baselines. Further, we conduct additional experiments on both datasets to illustrate how each component contributes to the final results.

\textbf{Abnormality classification.}
The comparison of multi-label abnormality classification in terms of AUC metric is shown in Table \ref{tab:classification_results}. Superior to the baselines that simply apply VGG-19\cite{simonyan2014very} or DenseNet-121\cite{huang2017densely} as backbone network with binary cross entropy loss, our method adds relationship constraint loss to exploit the relationships among abnormal medical terms, which outperforms baselines by 0.9\% of AUC on IU X-Ray dataset and 1.9\% on CX-CHR dataset. This demonstrates that our method is able to obtain useful visual features for correct abnormality classification via relationship constraint of medical terms. Its prominent capacity of detecting abnormalities is helpful for radiologists.

\begin{figure*}
   \centering
   \includegraphics[width=1.0\textwidth]{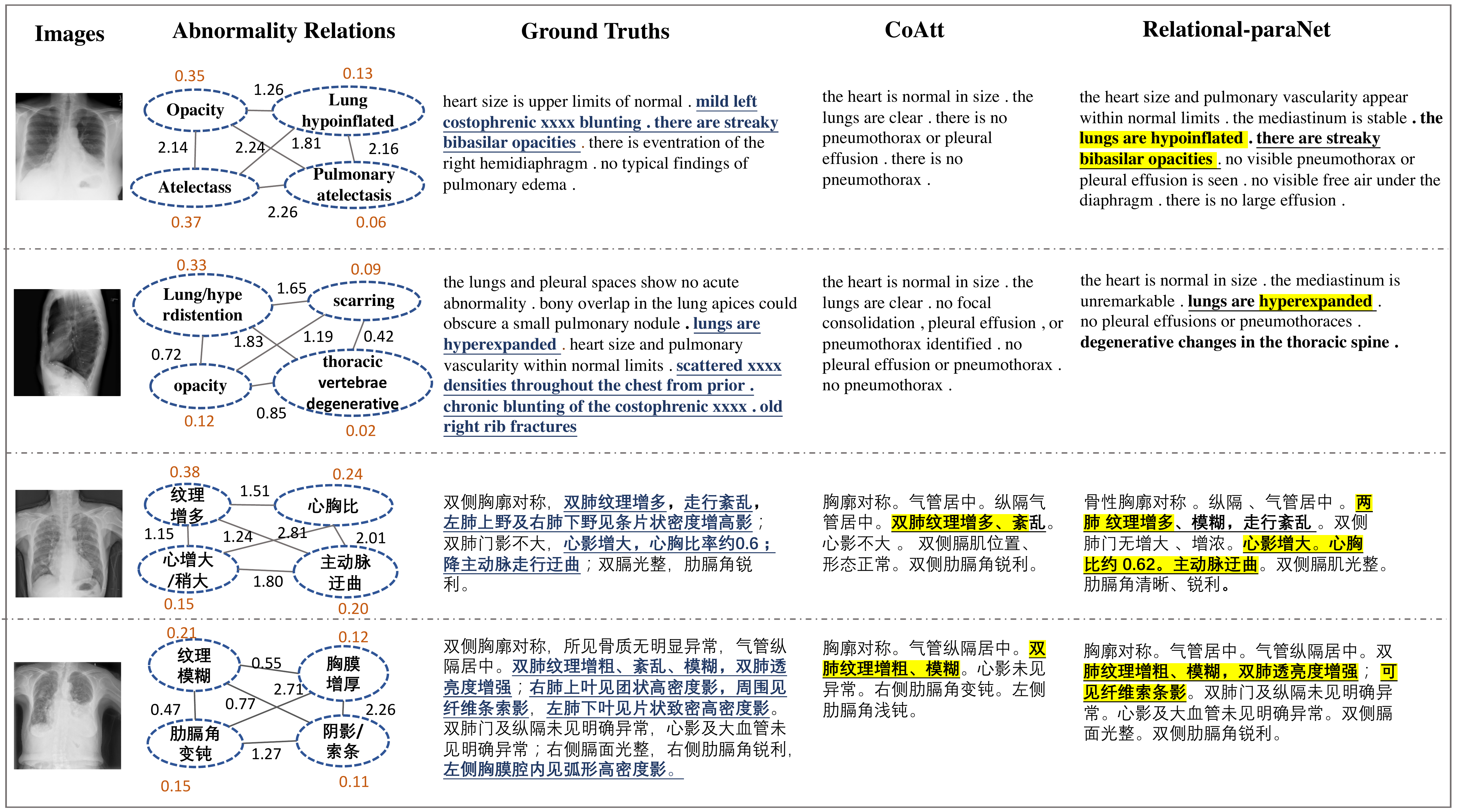}
   \caption{Visualization of results generated reports by CoAtt \cite{jing2017automatic} and our Relation-paraNet on both IU X-Ray (upper) and CX-CHR (lower). The underlined texts express the alignment between the generated text and ground truth reports. Bold texts indicate the correspondence of the retrieved text. The highlighted sections (yellow) illustrate the abnormal templates related to the abnormalities. In the abnormality relation graph, orange digits represent the classification scores. Black digits on the edges represent the relevance among abnormal medical terms, which is computed by Equation \ref{eq:relation_matrix_comput}.}
   \label{fig:visual}
\end{figure*}

\textbf{Automatic evaluation.} The automatic evaluation results under several metrics are shown in the Table \ref{tab:caption_results}. Obviously, our Relation-paraNet improves all metrics by large margins, demonstrating its effectiveness and extensiveness. CIDEr score represents inverse document frequency (IDF) of each vocabulary in the whole evaluated dataset, while BLEU-n matches n-grams within the evaluated sentences and ground truth sentences for each testing sample. In fact, improving the performance of BLEU-n with shorter n-gram (e.g., BLEU-1) is easier through generating common and seemingly correct words. On the contrary, CIDEr metric pays more attention to the critical but rare words under the consideration of IDF. Thus the CIDEr metric is more important on medical report composition, especially for the small dataset with unbalanced data.


On IU X-Ray dataset (i.e., a relatively small dataset including unbalanced data), the Relation-paraNet outperforms all baselines models (based on VGG-19 \cite{simonyan2014very}) on BLEU-1,2,3,4 scores, showing that the  Relational-Topic Encoder and the Adaptive Generator contribute to generating more professional reports.
HRGR-Agent  \cite{li2018hybrid} combines both the retrieval method and the generative model with reinforcement learning. Thus, it is reasonable that HRGR-Agent achieves the higher CIDEr score. However, our Ralation-paraNet still achieves the best CIDEr score, outperforming HRGR-Agent by 2.2\% due to the semantic consistency of abnormal medical terms and final reports encourage to generate rare abnormal descriptions.
In addition, when employing the DenseNet-121 \cite{huang2017densely} backbone as the recent work KERP \cite{li2019knowledge}, we can achieve much higher performance on all automatic metrics.

On CX-CHR dataset, Relation-paraNet achieves state-of-the-art performance. Our model improves CIDEr score by 0.399, ROUGE-L score by 5.7\%,  BLEU-1 score by 3.8\%, BLEU-2 score by 4.9\%, BLEU-3 score by 5.4\% and BLEU-4 score by 7.5\%, compared to KERP \cite{li2019knowledge}. Incorporating the semantic consistency of medical terms and considering abnormal sentences as templates, our Relation-paraNet encourages to produce reasonable and meaningful reports and achieves the best performance on automatic evaluation metrics.

To verify the effectiveness of our Relation-paraNet for medical report generation, we conduct extensive comparisons against several state-of-the-art visual captioning methods, such as AoA \cite{huang2019attention}, Transformer \cite{vaswani2017attention} and BUTD \cite{anderson2018bottom}. As we discussed in Sec. \ref{related_work}, the above attention-based methods are inapplicable to medical reports, which consists of multiple sentences or paragraphs.  And the results show that our Relation-paraNet outperforms these methods on all automatic evaluation metrics, demonstrating the superiority in retrieving abnormal sentences and generation normal ones. 

 \begin{table}
     \centering
     \caption{Human evaluation on both IU X-Ray and CX-CHR.}
     \resizebox{0.8\columnwidth}{!}{
     \begin{tabular}{c|c|c}
     \hline
     Dataset & Model & Hit(\%)  \\
     \hline
     \multirow{2}{*}{IU X-Ray} & CoAtt\cite{jing2017automatic} & 12.21  \\
     ~& Relation-paraNet & \textbf{35.51}   \\
     \hline
     \hline
     \multirow{2}{*}{CX-CHR} & CoAtt\cite{jing2017automatic} & 9.217 \\
     ~& Relation-paraNet & \textbf{67.254}  \\
     \hline
     \end{tabular}}
     \label{tab:human_evaluation}
     \vspace{-4mm}
 \end{table}

\textbf{Human evaluation.} Since Hybrid-agent \cite{li2018hybrid} and KERP \cite{li2019knowledge} have not released codes and models, our method is also compared with CoAtt \cite{jing2017automatic} in human evaluation, similar to them. We randomly select 100 generated medical reports of baseline model CoAtt \cite{jing2017automatic} and our methods in the test set. We invite five doctors to choose which one is more similar to ground truth under the consideration of abnormal findings, language fluency, and content coverage. We calculate the average preference percentages by excluding default choices (A default choice is provided in case of no or both reports are preferred), which is shown in Table \ref{tab:human_evaluation}. Obviously, our Relation-paraNet outperforms baseline a lot, indicating that it is capable of generating more accurate and reasonable reports.

\textbf{Conclusion.} Compared with Hybrid-agent \cite{li2018hybrid}, the proposed Relation-paraNet firstly preserves the superiority of bottom-up and top-down attention \cite{anderson2018bottom} for better semantic alignment between visual features and report descriptions. In addition, it is easier to generate abnormal descriptions by establishing the abnormal template datasets. Based on multi-layer and multi-head attention(Transformer) \cite{vaswani2017attention}, KERP \cite{li2019knowledge} built abnormality graph to learn relationships between abnormalities implicitly, the attention weights in which are lack of consistency and explanation \cite{jain2019attention}. In contrast, the Relation-paraNet contains a relationship constraint loss function to learn label co-occurrence dependencies directly. Since the popular visual captioning methods \cite{huang2019attention,vaswani2017attention,anderson2018bottom} is limited for paragraph generation, our Relation-paraNet derived the hierarchical structure from \cite{krause2017hierarchical} and achieved excellent performance in terms of paragraph generation.

\begin{figure*}
   \centering
   \includegraphics[width=1.0\textwidth]{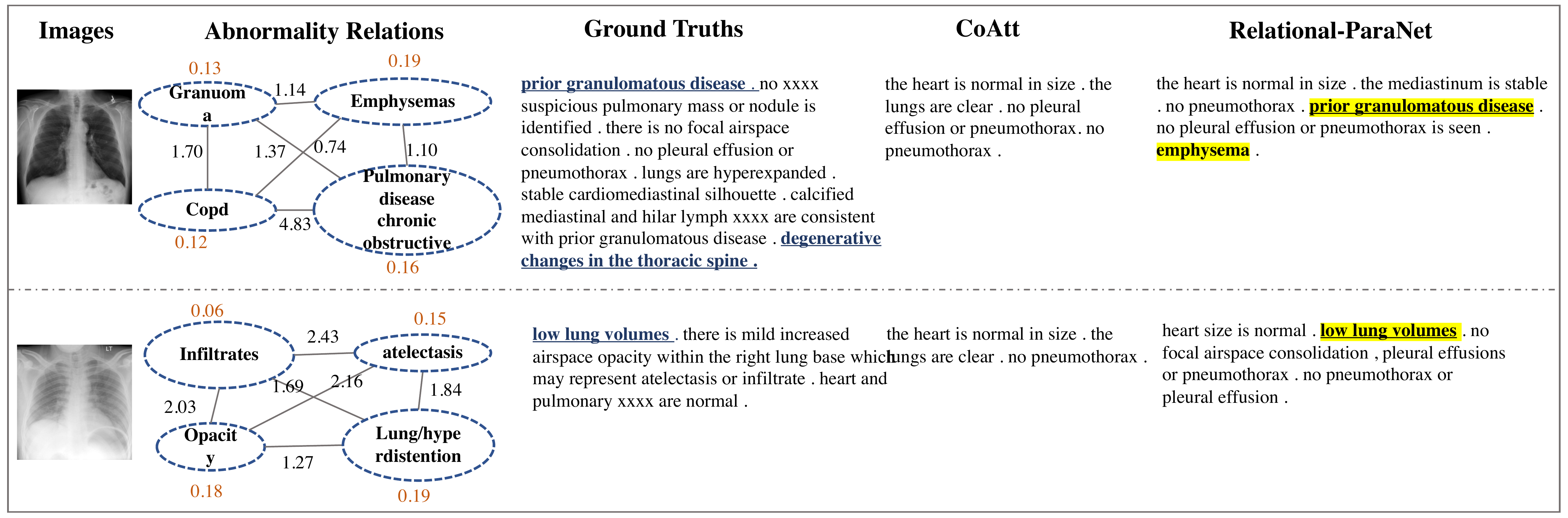}
   \caption{Another visualization of results generated by CoAtt\cite{jing2017automatic} and Relatoin-paraNet on IU X-Ray \cite{demner2015preparing}. The underlined text expresses alignment between the generated text and ground truth reports. Bold text indicates correspondence of the retrieved text. And the highlighted section illustrates abnormal templates are related to abnormalities. In the abnormality relation graph, orange digits represent the classification scores. Black digits on the edges represent the relevance among abnormal medical terms, which is computed by Equation \ref{eq:relation_matrix_comput}.}
   \label{fig:eng_visual2}
\end{figure*}

\begin{figure*}
   \centering
   \includegraphics[width=0.75\textwidth]{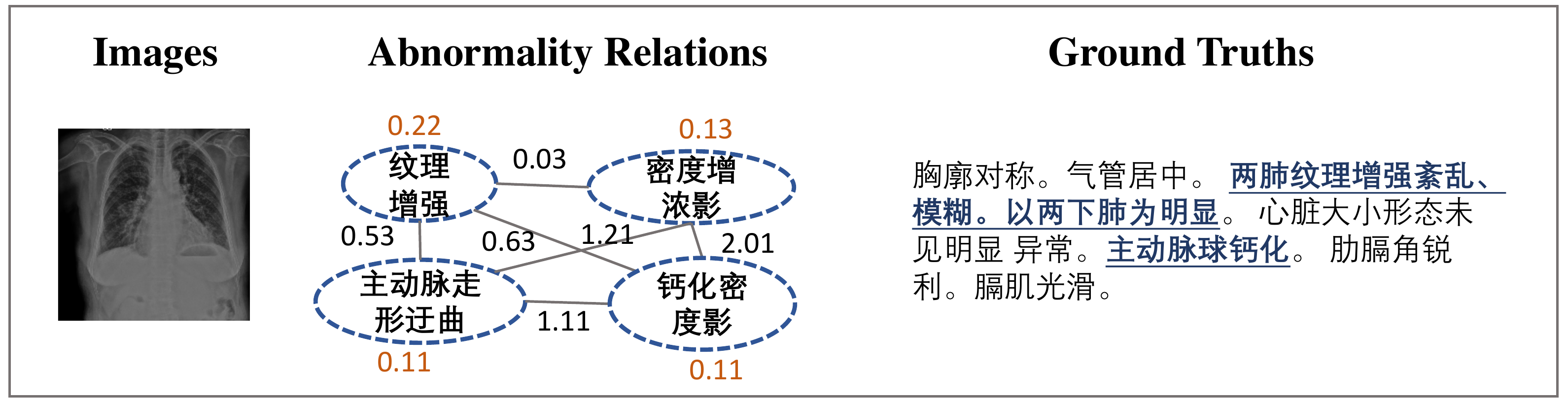}
   \caption{An image from CX-CHR dataset and its corresponding abnormality relations and ground truth report. In the abnormality relation graph, orange digits represent the classification scores. Black digits on the edges represent the relevance among abnormal medical terms, which is computed by Equation \ref{eq:relation_matrix_comput}.}
   \label{fig:ablation_image1}
\end{figure*}

\begin{figure*}
   \centering
   \includegraphics[width=1.0\textwidth]{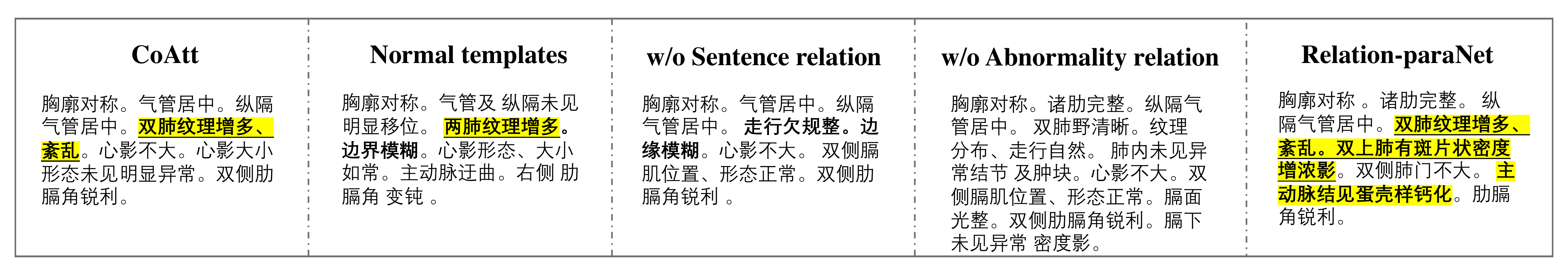}
   \caption{Qualitative ablation study results of the example shown in Figure \ref{fig:ablation_image1}. The underlined text expresses alignment between the generated text and ground truth reports. Bold text indicates correspondence of the retrieved text. And the highlighted section illustrates abnormal templates are related to abnormalities.}
   \label{fig:ablation_results1}
\end{figure*}

\begin{figure*}
  \centering
   \includegraphics[width=0.75\textwidth]{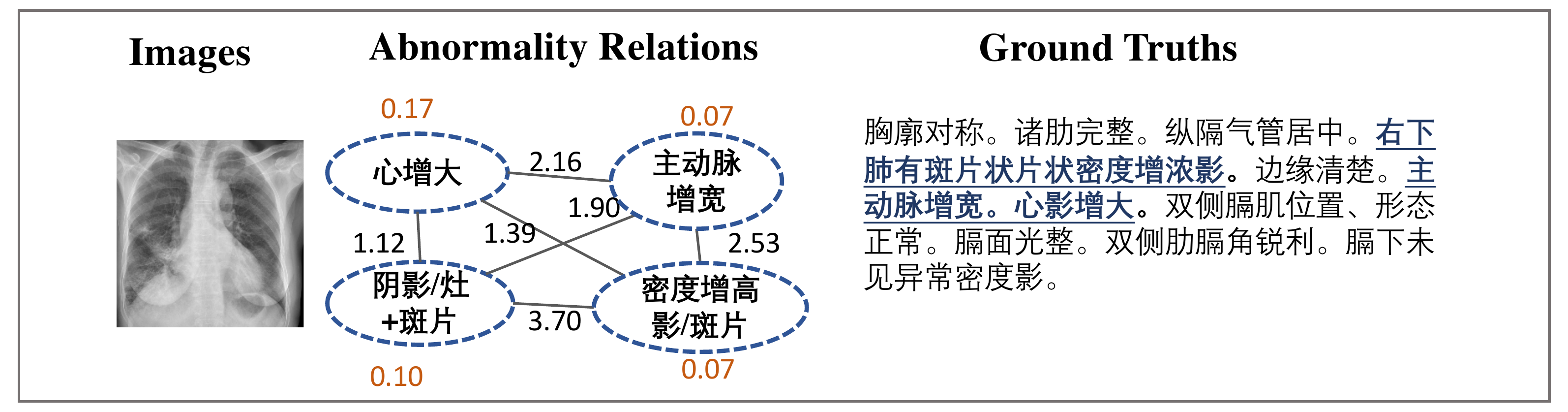}
   \caption{An image from CX-CHR dataset and its corresponding abnormality relations and ground truth report. In the abnormality relation graph, orange digits represent the classification scores. Black digits on the edges represent the relevance among abnormal medical terms, which is computed by Equation \ref{eq:relation_matrix_comput}.}
   \vspace{-4mm}
   \label{fig:ablation_image2}
\end{figure*}

\begin{figure*}
   \centering
   \includegraphics[width=1.0\textwidth]{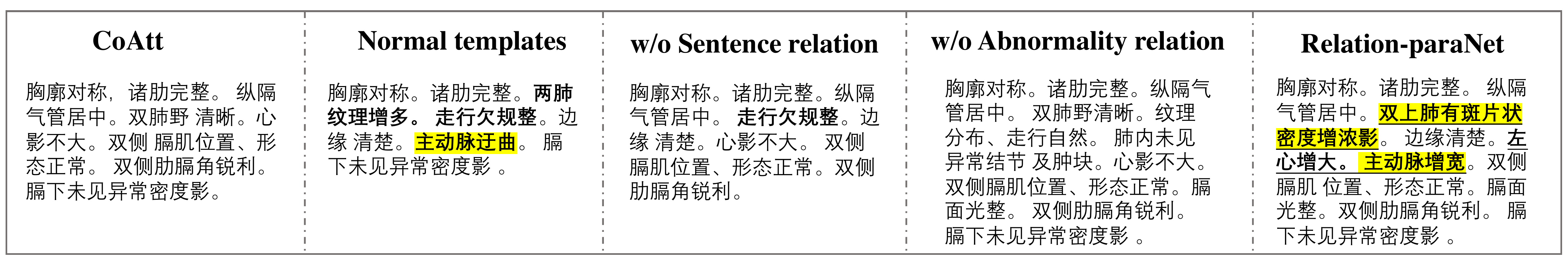}
   \caption{Qualitative ablation study results of the example shown in Figure \ref{fig:ablation_image2}. The underlined text expresses alignment between the generated text and ground truth reports. Bold text indicates correspondence of the retrieved text. And the highlighted section illustrates abnormal templates are related to abnormalities.}
   \label{fig:ablation_results2}
\end{figure*}

\subsection{Ablation Studies}
As shown in Sec. \ref{parameter_learning}, every term is necessary for the report generation task and any coefficient set to zero will lead to significant performance degradation. In this case, we paid more attention to illustrate the effectiveness of each module in our Relation-paraNet rather than adjusting these coefficients.
As manifested in Table \ref{tab:ablation_results}, all those modules contribute to better performance, compared with the final Relation-paraNet scores. Additional visualization results of ablation experiments are provided in Sec. \ref{qualitative_analysis}.

\textbf{Sentence relation.} Our model considers the sequence relationship between adjacent sentences, and encodes the last generated sentence information to guide in the next time step. This suggests once a sentence is generated correctly, it is useful to predict the next one. It encourages the Relation-paraNet to improve performance on all evaluation metrics, especially on BLEU-3 and BLEU-4.

\textbf{Normal templates vs. Abnormal templates.} The previous method HRGR-Agent \cite{li2018hybrid} first extracts normal sentences as templates and writes a new sentence for abnormal findings. In fact, it is easy for current generative model \cite{jing2017automatic} to fit common normal sentences. However, the rare abnormal sentences are more difficult to generate due to the unbalanced data. Instead, our Relation-paraNet regards the rare abnormal sentences as templates. The results show that this change can make the most improvement to our approach on the IU X-Ray dataset.

\textbf{Abnormality relation.} Our Relation-paraNet enforces the semantic consistency of medical terms to be incorporated into the final reports, as discussed in Sec. \ref{relational_abnormality}. Under the relationship constraint, more reasonable visual features are provided for the downstream modules, especially for the attention module. As shown in Figure  \ref{fig:visual}, templates retrieved by our Relation-paraNet are closely related to the abnormalities.
Moreover, we can observe our model improves BLEU-n metrics by a larger margin on the IU X-Ray dataset but improves CIDEr and ROUGH-L metrics on the CX-CHR dataset. This is due to the small-scale IU X-Ray dataset, the abnormality is necessary for the report composition. Therefore, the appropriate templates related to the abnormalities can promote BLEU-n metrics. On the other hand, without abnormality relation, our model can learn abnormal features along with downstream tasks adaptively due to the large-scale CX-CHR dataset. However, uncorrected abnormalities may cause unrelated templates. Therefore, the CIDEr and ROUGH-L metrics will be lower.

\subsection{Qualitative Analysis}
\label{qualitative_analysis}
Figure \ref{fig:visual} provides the visualized results generated by the baseline method and our Relation-paraNet on both IU X-Ray and CX-CHR dataset. It shows that CoAtt \cite{jing2017automatic} generates the most common sentences due to the dataset bias but gains not bad in terms of evaluation metrics. In the human evaluation, however, the performance of CoAtt is far worse than our Relation-paraNet (see Table \ref{tab:caption_results}), which is consistent with the visualization results. As shown in the first example, our system can produce a sentence ``there are streaky bibasilar opacities'' corresponding to the ground truth report. Specifically, our Relation-paraNet detects abnormality ``opacity'' first. Then the relation between ``opacity'' and ``there are streaky bibasilar opacities'' guides the system to choose this template. Moreover, the abnormality ``lung/hypoinflated'' is also related to ``opacity'', so our report contains ``the lungs are hypoinflated''. On CX-CHR dataset, our method also captures the critical topic accurately and generates more reasonable reports. Overall, unifying relational-topic driven retrieval and generation, our Relation-paraNet is able to write abnormal templates precisely and provide additional symptomatic references.

Additional visualization of results are shown in Figure \ref{fig:eng_visual2} - \ref{fig:ablation_results2}. In the abnormality relation graph, orange digits represent the classification scores. Black digits on the edges represent the relevance among abnormal medical terms. The underlined text expresses alignment between the generated text and ground truth reports. Bold text indicates the correspondence of the retrieved text. And the highlighted section illustrates inferred by abnormality relation and abnormal-template relation.

Figure \ref{fig:eng_visual2} provides another visualization of results generated by Relation-paraNet on IU X-Ray\cite{demner2015preparing}. Notice that our Relation-paraNet correctly produces ``prior granulomatous disease'' and ``low lung volumes'' in two examples,respectively.

Figure \ref{fig:ablation_image1}- \ref{fig:ablation_results2} present qualitative results of ablation study discussed in our paper. Obviously, our Relation-paraNet can retrieve templates according to the detected abnormalities exactly, compared with other models.

\section{Conclusion}

In this paper, we investigate incorporating hybrid knowledge co-reasoning upon the deep convolutional network to resolve the challenging medical report composition task. To enforce the semantic consistency of medical terms to be incorporated into the final reports and encourage the sentence generation for rare abnormal descriptions, we propose a novel Relation-paraNet that unifies template retrieval and sentence writing to handle both common and rare cases. Experiments on two medical report benchmarks demonstrate the superiority of our Relation-paraNet, which significantly outperforms the previous results in terms of all evaluation metrics as well as human evaluation.

In future work, we will explore the domain-specific learning difficulty issue resulting from unbalanced and insufficient datasets, build a more robust and reasonable relationship between abnormalities and corresponding abnormal templates and draw support from external medical knowledge and transfer the abstracted knowledge, to facilitate the vision-based paragraph generation.

\bibliographystyle{IEEEtran}
\bibliography{reference}

\vfill

\begin{IEEEbiography}[{\includegraphics[width=1in,height=1.25in,clip,keepaspectratio]{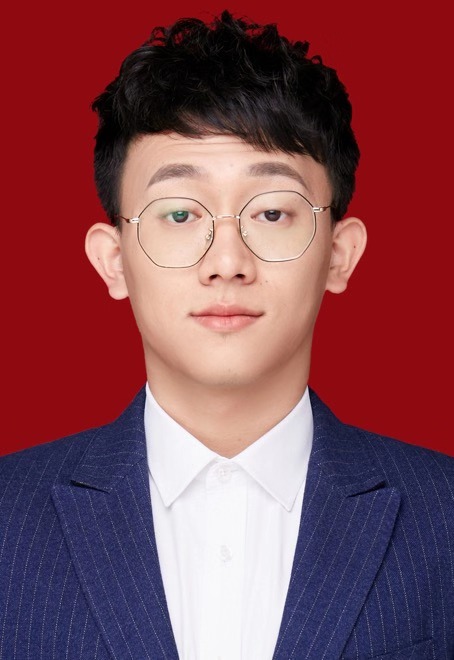}}]{Fuyu Wang}
 received his B.E. degree in Software Engineering from the
School of Data and Computer Science, Sun Yat-sen University,
Guangzhou, China, in 2018. He is currently
pursuing his  Ph.D. Degree in Computer Science
with the School of Data and Computer Science. His
current research interests include cross-modal learning and reasoning.
\end{IEEEbiography}

\begin{IEEEbiography}[{\includegraphics[width=1in,height=1.25in,clip,keepaspectratio]{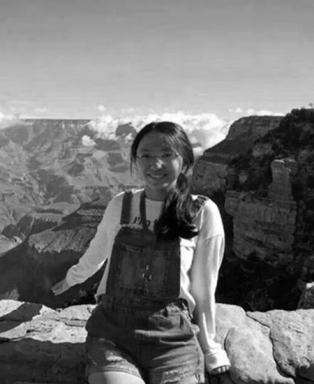}}]{Xiaodan Liang}
  Xiaodan Liang is currently an Associate Professor at Sun Yat-sen University. She was a postdoc researcher in the machine learning department at Carnegie Mellon University, working with Prof. Eric Xing, from 2016 to 2018. She received her PhD degree from Sun Yat-sen University in 2016, advised by Liang Lin. She has published several cutting-edge projects on human-related analysis, including human parsing, pedestrian detection and instance segmentation, 2D/3D human pose estimation and activity recognition.
\end{IEEEbiography}

\begin{IEEEbiography}[{\includegraphics[width=1in,height=1.25in,clip,keepaspectratio]{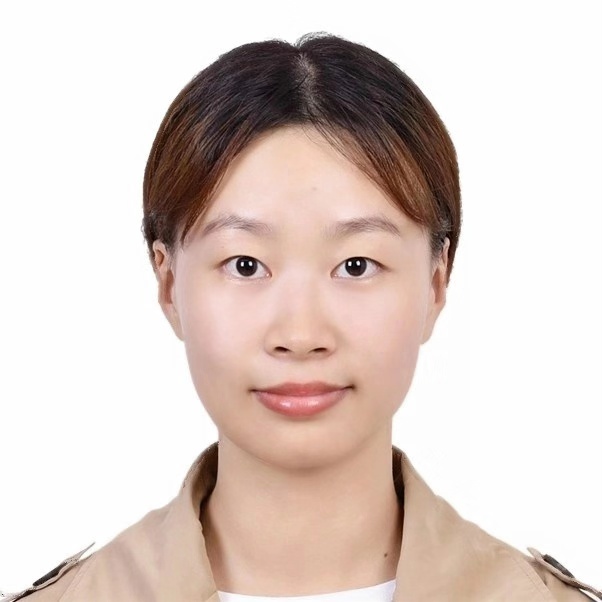}}]{Lin Xu}
 received his B.E. degree in Software Engineering from the
School of Data and Computer Science, Sun Yat-sen University,
Guangzhou, China, in 2018. He is currently
pursuing his  Ph.D. Degree in Computer Science
with the School of Data and Computer Science. His
current research interests include cross-modal learning and reasoning.
\end{IEEEbiography}

\begin{IEEEbiography}[{\includegraphics[width=1in,height=1.25in,clip,keepaspectratio]{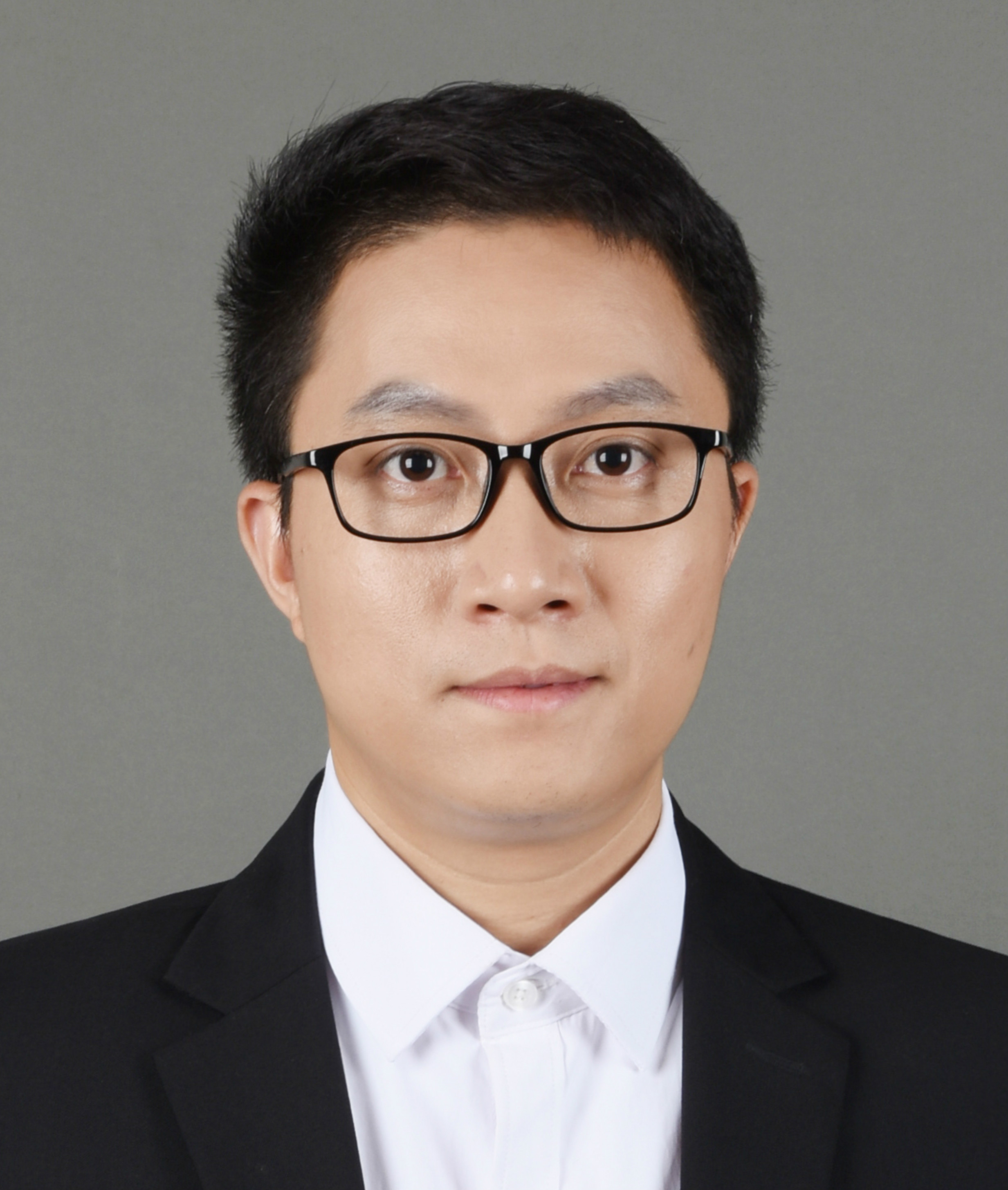}}]
{Liang Lin} is a Full Professor of computer science at Sun Yat-sen University. He has authored or co-authored more than 200 papers in top-tier academic journals and conferences with more than 12,000 citations. He is an associate editor of IEEE Trans. Human-Machine Systems and IET Computer Vision. He served as Area Chairs for numerous conferences such as CVPR, ICCV, and IJCAI. He is the recipient of numerous awards and honors including Wu Wen-Jun Artificial Intelligence Award, CISG Science and Technology Award, ICCV Best Paper Nomination in 2019, Annual Best Paper Award by Pattern Recognition (Elsevier) in 2018, Best Paper Dimond Award in IEEE ICME 2017, Google Faculty Award in 2012, and Hong Kong Scholars Award in 2014. He is a Fellow of IET.
\end{IEEEbiography}

\vfill

\end{document}